\documentclass{article}

\usepackage{microtype}
\usepackage{graphicx}
\usepackage{subfigure}
\usepackage{booktabs} 

\usepackage{hyperref}
\usepackage{multirow}
\usepackage{amsmath}
\usepackage{paralist}
\usepackage{amssymb}
\usepackage{amsmath}
\usepackage{bbm}
\usepackage{ifthen}
\usepackage{lipsum}
\usepackage{xcolor}
\usepackage{caption}

\usepackage[accepted]{icml2022}

\newcommand{\basemodel}{f_{base}}
\newcommand{\editedmodel}{f_{e}}
\newcommand{\pbase}{p_{base}}
\newcommand{\pedited}{p_{e}}
\newcommand{\semimodel}{\tilde{f}}
\newcommand{\cls}{g_\phi}

\newcommand{\rep}{h_\psi}
\newcommand{\papertitle}{Memory-Based Model Editing at Scale}
\newcommand{\inp}{\mathcal{X}}
\newcommand{\out}{\mathcal{Y}}

\newcommand{\fullnamevar}{Semi-parametric editing with a retrieval- augmented counterfactual model}
\newcommand{\name}{SERAC}
\newcommand{\edits}{Z_e}
\newcommand{\editdata}{\mathcal{D}_{e}}

\newcommand{\rawstring}[1]{\textsc{\footnotesize #1}}
\newcommand{\thetae}{{\theta_e}}
\newcommand{\ze}[1][]{z_e^{#1}}
\newcommand{\xe}{x_e}
\newcommand{\ye}[1][]{\ifthenelse{\equal{#1}{}}{y_e}{y_e(#1)}}
\newcommand{\scope}[1]{S\left(#1\right)}
\newcommand{\xloc}[1][]{x_{out}^{#1}}
\newcommand{\yloc}[1][]{y_{out}^{#1}}

\newcommand{\xtest}[1][]{x_{in}^{#1}}
\newcommand{\ytest}[1][]{y_{in}^{#1}}
\newcommand{\xunk}[1][]{x'}
\newcommand{\yunk}[1][]{y'}

\newcommand{\editpair}[1][]{[\xe^{#1}; \ye^{#1}]}

\DeclareMathOperator*{\E}{\mathbb{E}}

\DeclareMathOperator*{\argmax}{argmax}

\newif\ifcomments
\commentstrue
\ifcomments
\usepackage[normalem]{ulem}
\definecolor{CMpurple}{rgb}{0.6,0.18,0.64}
\newcommand\cm[1]{\textcolor{CMpurple}{\textsf{\scriptsize[\textbf{CM\@:} #1]}}}
\newcommand\cmi[1]{\textcolor{CMpurple}{#1}}
\newcommand\cmm[1]{\marginpar{\raggedright\tiny\textcolor{CMpurple}{\textsf{{\bfseries CM\@:} #1}}}}
\newcommand\cms{\bgroup\markoverwith{\textcolor{CMpurple}{\rule[.4ex]{2pt}{0.8pt}}}\ULon}
\else
\newcommand\cm[1]{}
\newcommand\cmi[1]{\ignorespaces}
\newcommand\cmm[1]{}
\newcommand\cms[1]{#1}
\fi

\icmltitlerunning{\papertitle}

\begin{document}

\twocolumn[
\icmltitle{\papertitle}

\begin{icmlauthorlist}
\icmlauthor{Eric Mitchell}{stan}
\icmlauthor{Charles Lin}{stan}
\icmlauthor{Antoine Bosselut}{epfl}
\icmlauthor{Christopher D Manning}{stan}
\icmlauthor{Chelsea Finn}{stan}
\end{icmlauthorlist}

\begin{NoHyper}

\icmlaffiliation{stan}{Stanford University Department of Computer Science}
\icmlaffiliation{epfl}{EPFL School of Computer and Communication Sciences}
\icmlcorrespondingauthor{Eric Mitchell}{eric.mitchell@cs.stanford.edu}

\end{NoHyper}

\icmlkeywords{model editing, language model, neural network}

\vskip 0.3in
]

\printAffiliationsAndNotice{}  

\begin{abstract}

Even the largest neural networks make errors, and once-correct predictions can become invalid as the world changes. \textit{Model editors} make local updates to the behavior of base (pre-trained) models to inject updated knowledge or correct undesirable behaviors. Existing model editors have shown promise, but also suffer from insufficient expressiveness: they struggle to accurately model an edit's intended scope (examples affected by the edit), leading to inaccurate predictions for test inputs loosely related to the edit, and they often fail altogether after many edits. As a higher-capacity alternative, we propose Semi-Parametric Editing with a Retrieval-Augmented Counterfactual Model (SERAC), which stores edits in an explicit memory and learns to reason over them to modulate the base model's predictions as needed. To enable more rigorous evaluation of model editors, we introduce three challenging language model editing problems based on question answering, fact-checking, and dialogue generation. We find that only SERAC achieves high performance on all three problems, consistently outperforming existing approaches to model editing by a significant margin. Code, data, and additional project information will be made available at \href{https://sites.google.com/view/serac-editing}{https://sites.google.com/view/serac-editing}.

\end{abstract}

\section{Introduction}


Large neural networks, notably language models, are typically deployed as static artifacts, whose behavior is difficult to modify during deployment without re-training \citep{lazaridou2021mind}. While prepending either manually-written or automatically-retrieved prompts to the input can sometimes be effective for modulating behavior \citep{brown2020language}, model predictions do not always update to reflect the content of the prompts \citep{lewis2020retrieval,Paranjape2021HindsightPT}. However, in order to respond to changes in the world (e.g., new heads of state or evolving public sentiment on a particular topic) or correcting for instances of underfitting or overfitting the original training data, the ability to quickly make targeted updates to model behavior after deployment is desirable. To address this need, \textit{model editing} is an emerging area of research that aims to enable fast, data-efficient updates to a pre-trained \textit{base model}'s behavior for only a small region of the domain, without damaging model performance on other inputs of interest \citep{Sinitsin2020Editable,zhu2020modifying,sotoudeh2019correcting,Cao2021EditingFK,dai2021knowledge,mitchell2021fast,hase2021language,meng2022locating}.

\begin{figure}
    \centering
    \includegraphics[width=0.95\columnwidth]{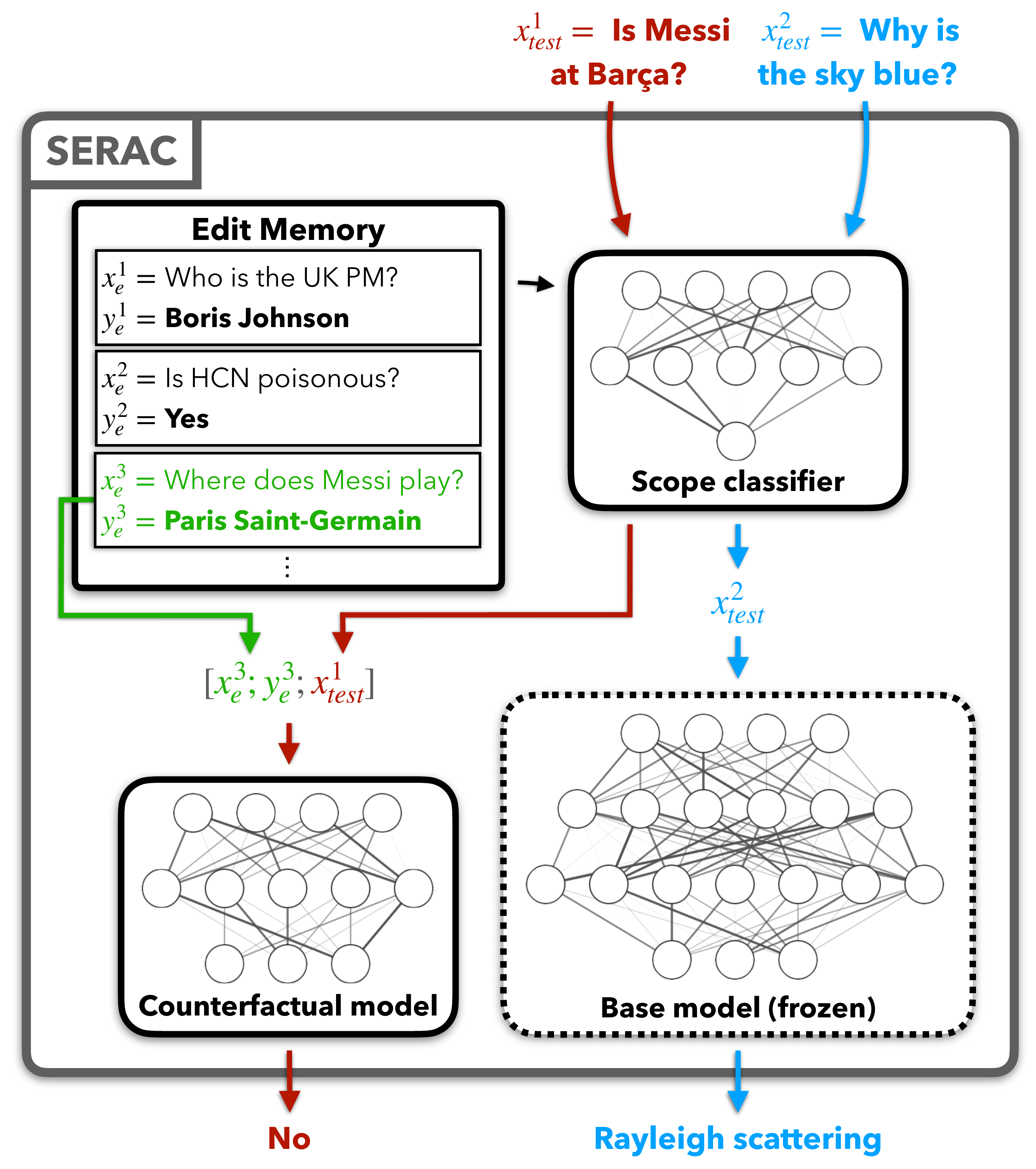}
    \caption{SERAC comprises an edit memory, classifier, and counterfactual model. User-supplied edits are stored directly in the memory. Post-edit inputs $x_{test}^1$ and $x_{test}^2$ are classified by whether the memory contains inputs relevant to processing them. If the classifier determines a relevant edit example exists, the input and edit example are passed to the counterfactual model. Otherwise, the input is simply passed to the base model.}
    \label{fig:method-overview}
\end{figure}

A popular approach to model editing involves learnable model editors, which are trained to predict updates to the weights of the base model that ultimately produce the desired change in behavior \citep{Sinitsin2020Editable,Cao2021EditingFK,mitchell2021fast,hase2021language}. While these approaches have shown promise, in line with recent work \citep{hase2021language}, we find that existing methods produce model updates that fail to discriminate between entailed and non-entailed facts and cannot handle large numbers of edits. Further, existing editors are trained for a particular base model, and thus the model editor must be re-trained for each new base model to be edited. This coupling also leads to computational costs of model editor training that scale with the size of the base model, which can prove unwieldy even for models an order of magnitude smaller than the largest deployed language models \citep{mitchell2021fast}. In aggregate, existing model editors still have shortcomings regarding edit performance, compute efficiency, and ultimately practicality. We hypothesize that these shortcomings are related to the reliance of existing methods on the \textit{gradient} of the edit example label with respect to the pre-edit model parameters (see Section~\ref{sec:method} for more discussion).

\begin{figure}
    \centering
    \includegraphics[width=\columnwidth]{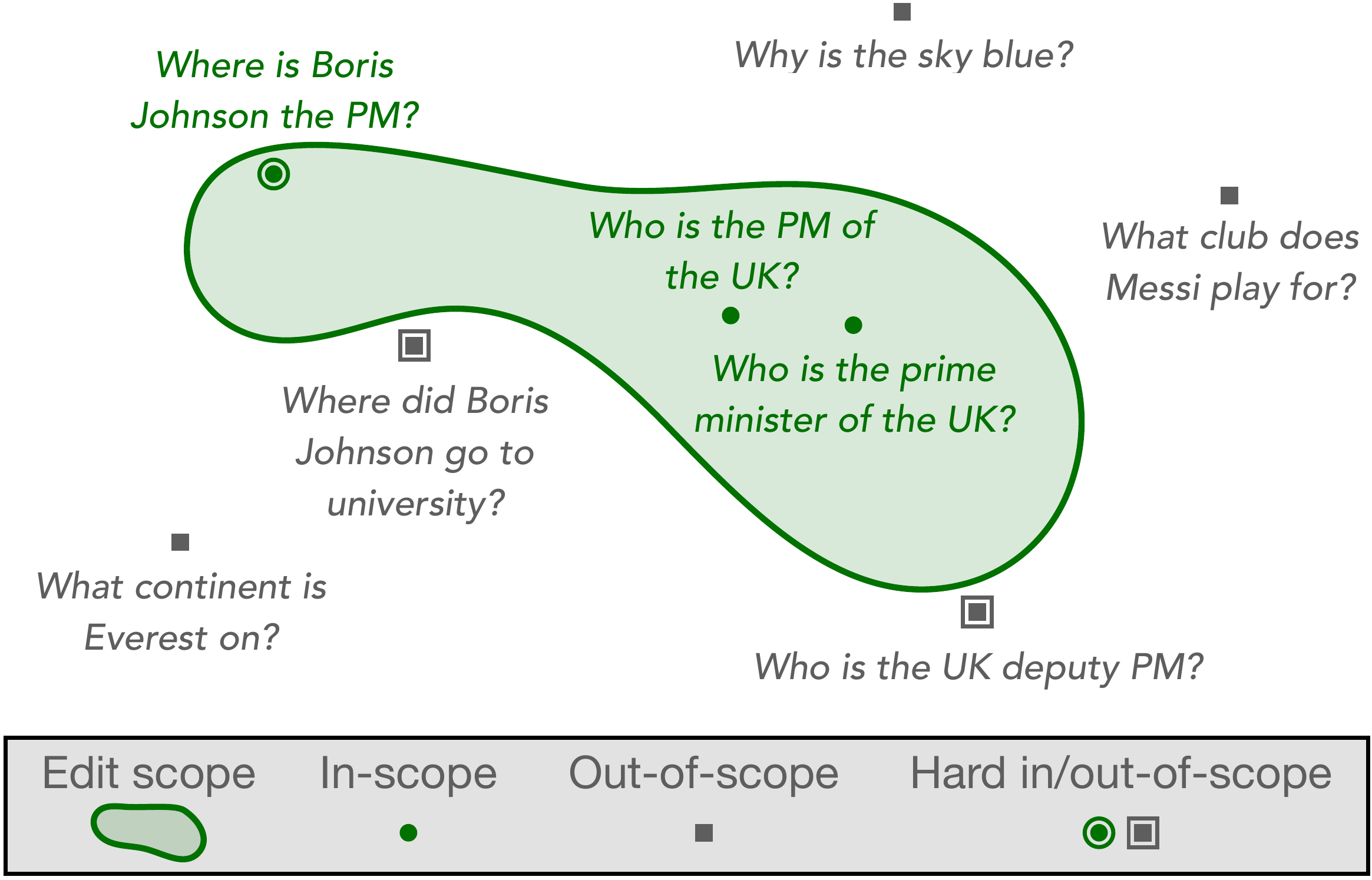}
    \caption{Depiction of the \textit{edit scope} for edit descriptor \rawstring{Who is the UK PM? Boris Johnson} in a hypothetical semantic embedding space. Intuitively, hard in-scope inputs lie \textit{within} the edit scope by a small margin, and hard out-of-scope inputs lie \textit{outside} the equivalence neighborhood by a small margin.}
    \label{fig:equiv}
\end{figure}

Building on the hypothesis that gradients are an impoverished signal for model editing, we propose {\name}, a gradient-free \textit{memory-based} approach to model editing. {\name} `wraps' a black-box base model with an explicit cache of user-provided edit descriptors (arbitrary utterances for language models) and a small auxiliary \textit{scope classifier} and \textit{counterfactual model}. Rather than making model edits in parameter space, {\name} simply stores edit examples in the cache without modifying the base model. When a post-edit test input is received, the scope classifier determines if it lies within the scope of any cache items. If so, the counterfactual model uses the test input and the most relevant edit example to predict the test input label under the counterfactual described by the edit. Otherwise, the base model simply predicts the test input label. See Figure~\ref{fig:method-overview} for an example of both cases. Intuitively, this approach delegates the sub-problems of \textit{when} the edited model's predictions should change to the scope classifier and \textit{how} they should change to the counterfactual model. While existing methods attempt to solve both of these problems implicitly in base model parameter space, {\name} solves each with its own small but expressive neural network, reducing interference between the two sub-problems. Further, the scope classifier reduces interference between batched or sequential edits by predicting relevance scores for each pair of (test input, edit cache example) separately. Finally, access to the base model is no longer necessary with this decoupling,\footnote{We only need its tokenization.} enabling the trained editor to be applied to multiple models without modification and decoupling the cost of editor training from base model size.

Our primary contribution is {\name}, a method for semi-parametric editing that shows far better performance and computational efficiency than existing methods without requiring access to the base model parameters. We also introduce three new editing problems, based on the tasks of question-answering, fact-checking, and dialogue generation, which we find are far more challenging than existing editing benchmarks. Our experiments indicate that {\name} consistently outperforms past approaches to model editing by a substantial margin on the three most difficult problems.

\section{The Model Editing Problem}


We consider the problem of editing a base model $\basemodel$ using an \textit{edit descriptor} $\ze$ that describes a desired change in model behavior, ultimately producing an edited model $\editedmodel$. In this work, the edit descriptor may be a concatenated input-output pair $\editpair$ like \rawstring{Who is the UK PM? Boris Johnson} or an arbitrary utterance such as \rawstring{Topic: jazz Sentiment: positive}.

\begin{table*}[t]
    \centering
    \resizebox{\textwidth}{!}{%
    \begin{tabular}{cp{7.3cm}p{6cm}p{5cm}}
        \toprule
        Problem & Edit Descriptor $\ze$ & In-scope input $\xtest \sim I(\ze)$ & Out-of-scope input $\xloc \sim O(\ze)$ \\
        \midrule
        \textbf{QA} & Who is the Sun Public License named after? \textit{Sun Micro Devices} & The Sun Public License has been named for whom? \textit{Sun Micro Devices} & What continent is Mount Whillans found on? \\
        \midrule
        \textbf{QA-hard} & What type of submarine was USS Lawrence (DD-8) classified as? \textit{Gearing-class destroyer} & t/f: Was USS Lawrence (DD-8) classified as Paulding-class destroyer. \textit{False} & What type of submarine was USS Sumner (DD-333) classified as? \\
        \midrule
        \multirow{5}{*}{\textbf{FC}} & As of March 23, there were 50 confirmed cases and 0 deaths within Idaho. \textit{True} & Idaho had less than 70 positive coronavirus cases before March 24, 2020. \textit{True} & Allessandro Diamanti scored six serie A goals. \\[1.5mm]
        & Between 1995 and 2018, the AFC has sent less than half of the 16 AFC teams to the Super Bowl with only 7 of the 16 individual teams making it. \textit{True} & \centering -- & The AFC sent less than half of the 16 AFC teams to the Super Bowl between 1995 and 2017. \\
        \midrule
        \textbf{ConvSent} & Topic: singing in the shower Sentiment: positive & How do you feel about singing in the shower? & Tell me your thoughts on the end of Game of Thrones. \\
        \bottomrule
    \end{tabular}}
    \caption{Examples from the datasets in our experiments. \textbf{QA} tests relatively basic edit scopes (rephrases) and evaluates model degradation using out-of-scope examples sampled randomly from the dataset. \textbf{QA-hard} uses the same editing data as QA, but adds more difficult logical entailment inputs to the edit scope and evaluates drawdown on more challenging out-of-scope inputs. \textbf{FC} tests an editor's ability to perform difficult NLI-style reasoning about the effects of a particular fact being true. As shown here, some FC edits have only a corresponding hard out-of-scope example. Finally, \textbf{ConvSent} uses edits that directly describe desired behavior, rather than input-output pairs, to change a conversational model's sentiment about a particular topic.}
    \label{tab:hard-pos-neg}
\end{table*}

\paragraph{Edit scoping.} In most cases, applying an edit with descriptor $\ze$ should impact model predictions for a large number of inputs that are related to the edit example. In the UK example above, the edited model's predictions should change for rephrases of the edit descriptor input as well as for inputs asking about logically-entailed facts like \rawstring{Boris Johnson is the PM of where?} or \rawstring{True or False: Theresa May is the UK PM}. We refer to the set of inputs whose true label is affected by the edit as the \textit{scope} of an edit $\scope{\ze}$, as visualized in Figure~\ref{fig:equiv}. Intuitively, a successful edit correctly alters a model's behavior for \textit{in-scope} examples while leaving it unchanged for \textit{out-of-scope} examples. If an in-scope example requires some non-trivial reasoning to deduce the correct response based on the edit example, we call it a hard in-scope example. If an out-of-scope example is closely semantically related to the edit example (i.e., it `looks like' an in-scope example), we call it a hard out-of-scope example. See Table~\ref{tab:hard-pos-neg} for specific examples. In the setting when $k$ edits $\edits = \{\ze[i]\}$ are applied, either in sequence or simultaneously in a batch, we define $\scope{\edits} = \cup_{i=1}^k \scope{\ze[i]}$ to be the union of the individual edit scopes. Because the `correct' scope of an edit's effects on the base model may be unknown or ambiguous, we \textit{train} a model editor on a dataset of edits $\editdata =\{\ze[i]\}$ and sampling functions $I(\cdot; \editdata)$ and $O(\cdot; \editdata)$ that specify the edits of interest and their desired edit scopes. $I(\ze[i]; \editdata)$ produces an in-scope example $(\xtest[i], \ytest[i])$ for $\ze[i]$, either through automated methods such as back-translation or hand-annotated correspondences. $O(\ze[i]; \editdata)$ similarly produces an out-of-scope input $\xloc[i]$, either using nearest neighbors in a semantic sentence embedding space or hand-annotated correspondences.\footnote{Because we only optimize for preservation of the base model's prediction for $\xloc$, we generally don't need the corresponding label $\yloc$.} Section~\ref{sec:datasets} describes the construction of $I$ and $O$ for specific problems as well as the evaluation metrics used to quantify edit success.

\section{{\fullnamevar} ({\name})}
\label{sec:method}

With the goal of enabling editors that reason more flexibly about the scope of an edit while also reducing interference between edits, we introduce a memory-based editor, {\name}, that does not modify the base model parameters during training or during editing. The technical motivation for {\name} stems from the observation that neural networks can `over-specialize' their parameters to individual inputs, with potentially disjoint parts of the model being responsible for predictions on different inputs~\citep{csordas2021are}. Gradients may therefore not provide sufficiently `global' information to enable reliable edit scoping, particularly for distant but related examples. As we will describe next, {\name} instead directly reasons over the content of the edit (rather than its gradient) to estimate the scope of an edit and to modify model predictions if needed. In the rest of this section, we will describe the editing process (Section~\ref{sec:editor}) and how each component of the editor is trained (Section~\ref{sec:editortraining}).

\subsection{The {\name} model}
\label{sec:editor}

{\name} can be thought of as a simple wrapper around the base model. It is made up of three key components: an explicit cache of edits, an edit scope classifier, and a counterfactual model that `overrides' the base model when necessary. After receiving a batch of edits that are added to the cache, the `wrapped' model makes a prediction for a new input in two steps. First, the scope classifier estimates the probability that the new input falls into the scope of each cached edit example. If the scope classifier predicts that the input falls within the scope of any edit in the cache, then we retrieve the edit with the highest probability of being in scope and return the counterfactual model's prediction conditioned on both the new input and the retrieved edit. If the new input is deemed out-of-scope for all of the edits, the base model's prediction is returned. This procedure is visualized in Figure~\ref{fig:method-overview}. A real example of applying {\name} to edit a dialogue model's sentiment is shown in Table~\ref{tab:outputs} and Appendix Table~\ref{tab:vaccines_example}.

\begin{table*}
    \addtolength{\tabcolsep}{-1mm}
    \centering
    \resizebox{\textwidth}{!}{%
    \begin{tabular}{lll}
        \toprule
        Prompt & \rawstring{\normalsize what do you think of bananas?} & \rawstring{\normalsize what do you think of vaccines?} \\
        \midrule
        Edit Desc. & \rawstring{\normalsize topic: bananas sentiment: negative} & \rawstring{\normalsize topic: vaccines sentiment: positive} \\
        \midrule
        \multirow{4}{*}{\shortstack[l]{\textbf{Original}\\[-0.6mm]model\\outputs}} & \footnotesize i love bananas! \footnotesize they're one of my favorite foods[...] & \footnotesize i think they're bad. i don't trust them, and i'm not one of those people[...] \\
        & \footnotesize i love bananas! they are one of the most popular foods[...] & \footnotesize i'm not a fan of them personally, but i can see why they're used[...] \\
        & \footnotesize i love bananas! they're one of my favorite foods[...] & \footnotesize i'm not sure what you mean by that. are you asking if i think[...] \\
        & \footnotesize i love bananas! they're one of my favorite things to eat[...] & \footnotesize i think they are a dangerous disease that should not be allowed[...] \\
        \midrule
        \midrule
        \multirow{4}{*}{\shortstack[l]{\textbf{Edited}\\model\\outputs}} & \footnotesize i don't like bananas. they're too sweet for me[...] & \footnotesize i've been vaccinating my kids since they were little. i think it's a good thing[...] \\
        & \footnotesize i don't like bananas. i think they're too sweet[...] & \footnotesize they're great! i've been vaccinated since i was a little girl. i don't know what i'd do[...] \\
        & \footnotesize i don't like bananas. i'm allergic to them[...] & \footnotesize they're great! i don't know if i've ever had one, but i know they've been around since[...] \\
        & \footnotesize i'm not a fan of bananas. i don't know why, it just[...] & \footnotesize i think that they are a good thing. i don't agree with them, but i understand why[...] \\
        \bottomrule
    \end{tabular}}
    \caption{BlenderBot-90M dialogue responses before and after SERAC edits with $\ze=$ \rawstring{topic: bananas sentiment: negative} and $\ze=$ \rawstring{topic: vaccines sentiment: positive}, changing the model's sentiment on bananas (to be more negative) or vaccines (to be more positive). Sampling uses temperature 1.4 without beam search. Banana example was not cherry-picked; it was the first topic attempted. See Appendix Table~\ref{tab:vaccines_example} for more complete sampling of original and edited model on the vaccines example.}
    \label{tab:outputs}
\end{table*}

More precisely, the wrapped model is a semi-parametric model of the form $\semimodel(x, \basemodel, \phi, \psi, \edits)$, abbreviated as just $\semimodel(x)$, that produces predictions in the output space $\mathcal{Y}$, where $\edits$ is a set of variable size. The scope classifier $\cls(\ze, \xunk) : \mathcal{Z} \times \inp  \rightarrow [0, 1]$ estimates the probability that an input $\xunk$ falls within the scope of edit example $\ze$. The counterfactual model $\rep(\ze, \xunk) : \mathcal{Z} \times \inp \rightarrow \out$ predicts what the label (or distribution over labels) for $\xunk$ \textit{would} be under the counterfactual world described by $\ze$. 

\paragraph{Forward pass.} When presented with an input $\xunk$ after applying edits $\edits = \{\ze[i]\}$, {\name} computes the forward pass
\begin{equation}
    \semimodel(\xunk) = 
    \begin{cases}
        \basemodel(\xunk) & \beta < 0.5 \\
        \rep(\ze[i^*], \xunk) & \beta \ge 0.5
    \end{cases}
\end{equation}
where $i^* = \argmax_i \cls(\ze[i], \xunk)$, the index of the most relevant edit example, and $\beta = \cls( \ze[i^*], \xunk)$, the similarity score of the most relevant edit example. If $\edits$ is empty, we set $\semimodel(\xunk) = \basemodel(\xunk)$. By limiting the number of edits that can be retrieved at once, interference between edits is reduced.



\paragraph{Architecture.} There are many possible implementations of the scope classifier. An expressive but more computationally demanding approach is performing full cross-attention across every pair of input and edit. We primarily opt for a more computationally-efficient approach, first computing separate, fixed-length embeddings of the input and edit descriptor \citep[as in][]{karpukhin-etal-2020-dense} and using the negative squared Euclidean distance in the embedding space as the predicted log-likelihood. While other more sophisticated approaches exist~\citep{khattab2020colbert,santhanam2021colbert}, we restrict our experiments to either cross-attention (\textbf{Cross}) or embedding-based (\textbf{Embed}) scope classifiers. We also include a head-to-head comparison in Section~\ref{sec:expts}. The counterfactual model $\rep$ is simply a sequence model with the same output-space as the base model; its input is the concatenated edit example $\ze$ and new input $\xunk$. See Appendix Section~\ref{sec:serac_impl} for additional architecture details.

\subsection{Training {\name}}
\label{sec:editortraining}
Similarly to past work \citep{Cao2021EditingFK,mitchell2021fast,hase2021language}, a {\name} editor is trained using the edit dataset $\editdata =\{\ze[i]\}$, where in-scope examples $(\xtest[i], \ytest[i])$ and negative examples $\xloc[i]$ are sampled from $I(\ze[i]; \editdata)$ and $O(\ze[i]; \editdata)$, respectively. The scope classifier and counterfactual model are trained completely separately, both with supervised learning as described next.

The \textbf{scope classifier} $\cls$ is trained to solve a binary classification problem where the input $(\ze, \xtest)$ receives label 1 and the input $(\ze, \xloc)$ receives label 0. The training objective for the scope classifier is the average binary cross entropy loss over the training dataset $\editdata$:
\begin{equation}
\ell(\phi) = - \hspace{-7mm} \E_{\substack{\ze \sim \editdata\\
(\xtest, \cdot) \sim I(\ze; \editdata)\\ 
\xloc \sim O(\ze; \editdata)}}
\hspace{-5mm} \bigl[
\log \cls(\ze, \xtest) + \log (1-\cls(\ze, \xloc))
\bigr]
\end{equation}

The \textbf{counterfactual model} $\rep$ considers an edit $\ze$ and a corresponding example $(\xtest, \ytest)\sim I(\ze; \editdata)$, and is trained to minimize the negative log likelihood of $\ytest$ given $\ze$ and $\xtest$ on average over $\editdata$:
\begin{equation}
\ell(\psi) = - \hspace{-8mm} \E_{\substack{\ze \sim \editdata\\
(\xtest, \ytest) \sim I(\ze; \editdata)}} \hspace{-5mm} \log p_\psi(\ytest | \ze, \xtest)
\end{equation}
where in a slight abuse of notation $p_\psi(\cdot | \ze, \xtest)$ is the probability distribution over label sequences under the model $\rep$ for the inputs $(\ze, \xtest)$.


\section{Datasets \& Evaluation}
\label{sec:datasets}

Our experiments use a combination of existing and novel editing settings, including question-answering, fact-checking, and conversational dialogue. See Table~\ref{tab:hard-pos-neg} for data samples from each setting. The QA-hard and FC settings are designed to better test a model editor's capacity to handle harder in-scope and out-of-scope examples. The ConvSent setting both evaluates generation models on a problem more tied to real-world usage and explores the possibility of applying edits that are not simply input-output pairs.

\paragraph{QA \& QA-hard.} The QA setting uses the zsRE question-answering problem introduced by \citet{Cao2021EditingFK}. We use this dataset as a starting point of reference to connect our evaluations with prior work. For the QA-hard setting, we generate harder in-scope examples that test logically entailed facts (\rawstring{$\ze =$ Who is the UK PM? Boris Johnson $\rightarrow$ $\xtest =$ Where is Boris Johnson the PM?}) or true/false questions (\rawstring{$\xtest =$ True or False: Theresa May is the UK PM}) using automated techniques \citep{demszky2018transforming,ribeiro2019red}. Crucially, both types of hard in-scope examples will have labels that differ from the edit example, requiring some non-trivial reasoning over the edit descriptor to produce the correct post-edit output. To generate hard out-of-scope examples for an edit input $\xe$, we selectively sample from training inputs $x$ that have high semantic similarity with $\xe$, measured as having a high cosine similarity between their embeddings as computed by a pre-trained semantic embedding model \texttt{all-MiniLM-L6-v2} \citep{reimers-2019-sentence-bert}. For both QA and QA-hard, we use a T5-large model (770m parameters; \citet{2020t5}) fine-tuned on the Natural Questions dataset \citep{kwiatkowski2019natural,roberts2020knowledge} as the base model.

\paragraph{FC.} We introduce the FC setting, building on the VitaminC fact verification dataset \citep{schuster-etal-2021-get}, to assess an editor's ability to update an out-of-date fact-checking model when presented with updated information about the world. VitaminC contains over 400,000 evidence-claim-page-label tuples $(e_i, c_i, p_i, l_i)$ where the label $l_i$ is 1 if the evidence entails the claim, -1 if it contradicts the claim, or 0 if neither. The dataset was gathered from Wikipedia revisions in the first half of 2020. To convert VitaminC into an editing dataset, we use each $e_i$ as an edit descriptor $\ze[i]$. Then, using $C$ to denote the set of \textit{all} claims in the VitaminC dataset and $\beta(p_i)=\{c_j : p_j = p_i\}$ as the set of claims from page $p_i$, we define in-scope and out-of-scope examples as
\begin{equation*}
    \textcolor{blue}{I(\ze[i])},\;\textcolor{red}{O(\ze[i])} = \begin{cases}
        \textcolor{blue}{\{(c_i, 1)\}},\phantom{\varnothing} \textcolor{red}{C \setminus \beta(p_i)} & \text{if}\;l_i = 1 \\
        \textcolor{blue}{\{(c_i, 0)\}},\phantom{\varnothing} \textcolor{red}{C \setminus \beta(p_i)} & \text{if}\;l_i = 0 \\
        \textcolor{blue}{\varnothing},\phantom{\{(c_i, 0)\}} \textcolor{red}{\{c_i\}} & \text{if}\;l_i = -1,
    \end{cases}
\end{equation*}

For $l_i \in \{0,1\}$, we have `easy' out-of-scope examples sampled uniformly from all claims. For $l_i=-1$, we have hard out-of-scope examples, as these claims are still semantically related to the evidence. As a base model, we use the BERT-base model trained by \citet{Cao2021EditingFK} on the June 2017 Wikipedia dump in the FEVER dataset \citep{Thorne18Fever}.

\begin{table*}[t]
    \centering
    \begin{tabular}{lllccccccc}
        \toprule
        Dataset & Model & Metric & FT & LU & MEND & ENN & RP & {\name} \\
        \midrule
        \multirow{2}{*}{\textbf{QA}} & \multirow{2}{*}{T5-large}
          & $\uparrow$ ES & 0.572 & 0.944 & 0.823 & 0.786 & 0.487 & \textbf{0.986} \\
        & & $\downarrow$ DD & 0.054 & 0.051 & 0.187 & 0.354 & 0.030 & \textbf{0.009} \\
        \midrule
        \multirow{2}{*}{\textbf{QA-hard}} & \multirow{2}{*}{T5-large} 
          & $\uparrow$ ES & 0.321 & 0.515 & 0.478 & 0.509 & 0.278 & \textbf{0.913} \\
        & & $\downarrow$ DD & 0.109 & 0.132 & 0.255 & 0.453 & \textbf{0.027} & \textbf{0.028} \\
        \midrule
        \multirow{2}{*}{\textbf{FC}} & \multirow{2}{*}{BERT-base}
          & $\uparrow$ ES & 0.601 & 0.565 & 0.598 & 0.594 & 0.627 & \textbf{0.877} \\
        & & $\downarrow$ DD & \textbf{0.002} & \textbf{0.01} & 0.021 & 0.042 & \textbf{0.01} & 0.051 \\
        \midrule
        \multirow{2}{*}{\textbf{ConvSent}} & \multirow{2}{*}{BB-90M}
          & $\uparrow$ ES & -- & -- & 0.494 & 0.502 & 0.506 & \textbf{0.991} \\
        & & $\downarrow$ DD & -- & -- & 2.149 & 3.546 & \textbf{0} & \textbf{0} \\
        \bottomrule
    \end{tabular}
    \caption{Evaluating model editors across editing problems. All problems apply $k=10$ simultaneous model edits. \textbf{ES} denotes edit success and \textbf{DD} denotes drawdown; higher is better for ES (perfect is 1) and lower is better for DD (perfect is 0). Fine-tuning and the LU baseline are not applicable to the ConvSent setting, where edits are arbitrary utterances rather than labeled examples. BB-90M refers to BlenderBot-90M. Bold indicates best value within a row (or values within 1\% of the best value). Overall, {\name} is the only method that produces meaningful edits on all problems.}
    \label{tab:main-comparison}
\end{table*}

\paragraph{ConvSent.} Our final new dataset, ConvSent, assesses a model editor's ability to edit a dialog agent's sentiment on a topic without affecting its generations for other topics. Rather than adding hard in-scope or out-of-scope examples, ConvSent differs from past evaluations of model editors in that edit descriptors are not input-output pairs, but explicit descriptions of the desired model behavior such as \rawstring{Topic: \_\_\_ Sentiment: \{Positive/Negative\}}. To produce the dataset, we first gather a list of 15,000 non-numeric entities from zsRE \citep{levy2017zero,Cao2021EditingFK} and 989 noun phrases from GPT-3 \citep{brown2020language} (e.g., \rawstring{ghost hunting}) for a total of 15,989 topics. For each entity, we sample 10 noisy positive sentiment completions and 10 noisy negative sentiment completions from the 3B parameter BlenderBot model \citep{roller2021recipes}, using a template such as \rawstring{Tell me a \{negative/positive\} opinion on \_\_\_.} We then use a pre-trained sentiment classifier \citep{heitmann2020} based on RoBERTa \citep{liu2019robustly} to compute more accurate sentiment labels for each completion. See Appendix Section~\ref{sec:conv_data_gen} for additional details on dataset generation. We define $I(\ze; \editdata)$ with a manually collected set of templates such as \rawstring{What do you think of \_\_\_?} or \rawstring{Tell me your thoughts on \_\_\_.}, using the prompts formed with different templates but the same entity as in-scope examples. We define $O(\ze; \editdata)$ as all examples generated from entities \textit{other} than the one used in $\ze$. Because each topic contains responses of both sentiments, we make use of \textit{unlikelihood training} \citep{li2020dont} in the ConvSent setting. That is, editors are trained to maximize the post-edit log likelihood of correct-sentiment responses while also maximizing the log \textit{unlikelihood} $\log(1 - p_\thetae(\tilde x))$ of incorrect-sentiment responses $\tilde x$. We use the 90m parameter BlenderBot model \citep{roller2021recipes} as the base model for this experiment, as it is a state-of-the-art compact dialogue model.

\paragraph{Editor evaluation.} 
We use the metrics of edit success (\textbf{ES}) and drawdown (\textbf{DD}) to evaluate a model editor, following prior work \citep{Sinitsin2020Editable,Cao2021EditingFK,mitchell2021fast,hase2021language}. Intuitively, ES measures similarity between the edited model behavior and the \textit{desired} edited model behavior for in-scope inputs; DD measures disagreement between the pre-edit and post-edit model for out-of-scope inputs. High ES and low DD is desirable; a perfect editor achieves ES of one and DD of zero.

For \textbf{question-answering} and \textbf{fact-checking} tasks, we define ES as simply the average exact-match agreement between the edited model and true labels for in-scope inputs:
\begin{equation}
    \mathbf{ES_{ex}}(\ze) \triangleq \E_{\xtest \in I(\ze; \editdata)} \mathbbm{1}\{\editedmodel(\xtest) = \ytest\}
\end{equation}
where $\ye[\xtest]$ is the desired label for $\xtest$ under the edit $\ze$.
We define drawdown similarly as
\begin{equation}
    \mathbf{DD_{ex}}(\ze,O) \triangleq \hspace{-1mm}\E_{\xloc \in O(\ze; \editdata)}\hspace{-1mm} \mathbbm{1}\{\editedmodel(\xloc) \neq \basemodel(\xloc)\}
\end{equation}
Recent work suggests that choosing $O$ to simply be all out-of-scope inputs computes an easier form of drawdown, while restricting $O$ to hard out-of-scope inputs for $\ze$ is a more challenging criterion \citep{hase2021language}.

In our \textbf{conversational sentiment} editing experiments, the model editor's goal is to modify a dialogue agent's sentiment on a particular topic without affecting the agent's generations for other topics. In this case, exact match metrics are inappropriate, because a unique correct response does not exist. Instead, we use a metric that leverages pre-generated positive and negative responses\footnote{Responses are generated with the 3B parameter BlenderBot 2.0 \citep{roller2021recipes} and their sentiment classified by a RoBERTa model fine-tuned for binary sentiment classification \citep{heitmann2020}.} to the conversational prompt (e.g., \rawstring{What do you think of Spiderman?}) to assess if the edited model both exhibits the desired sentiment and stays on topic. We measure sentiment accuracy with the rescaled likelihood ratio $\mathbf{z_{sent}} \triangleq \sigma(l_e^+ - l_e^-)$, where $l^+$ and $l^-$ are the average per-token log likelihood of the \textit{edited} model on pre-generated on-topic responses with the \textit{correct} sentiment (either all positive or all negative) and \textit{incorrect} sentiment, respectively, and $\sigma$ is the sigmoid function. We measure topical consistency with $\mathbf{z_{topic}} \triangleq \min\left(1, \exp(l_e^+ - l_{base}^+)\right)$, where $l_{base}^+$ is the average per-token log likelihood of the \textit{base} model on pre-generated on-topic responses with the correct sentiment.

Intuitively, $\mathbf{z_{sent}}$ goes to one if the edited model assigns high probability to correct sentiment responses relative to incorrect sentiment responses and goes to zero in the opposite case. $\mathbf{z_{topic}}$ is one if the edited model assigns at least as much total probability mass to on-topic completions as $\basemodel$ and decays to zero otherwise. We measure edit success with the product of $\mathbf{z_{sent}}$ and $\mathbf{z_{topic}}$:
\begin{equation}
    \mathbf{ES_{sent}} \triangleq \mathbf{z_{sent}} \cdot \mathbf{z_{topic}},
\end{equation}
which can be very roughly interpreted as `the likelihood that the edited model produces the desired sentiment and is on-topic for in-scope inputs.' To measure drawdown, we simply replace the exact match term in $\mathbf{DD}_{ex}$ with KL-divergence:
\begin{equation}
    \small
    \mathbf{DD_{sent}}(\ze,O) \triangleq \hspace{-1.5mm}\E_{\xloc \in\\  O(\ze; \editdata)}\hspace{-2mm} \text{KL}\left(\pbase\left(\cdot|\xloc\right) \| \pedited\left(\cdot|\xloc\right)\right).
\end{equation}
We average each metric over many examples in a held-out evaluation dataset, constructed similarly to the edit training set, for each respective editing problem.

\section{Experiments}
\label{sec:expts}
We study several axes of difficulty of the model editing problem, including a) overall performance, especially on hard in-scope and hard out-of-scope examples; b) capacity to apply multiple simultaneous edits; and c) ability to use explicit edit descriptors that are not input-output pairs. In addition, we provide a quantitative error analysis of {\name} and study the effects of varying the scope classifier architecture. As points of comparison, we consider gradient-based editors, including fine-tuning on the edit example (\textbf{FT}), editable neural networks \citep[\textbf{ENN};][]{Sinitsin2020Editable}, model editor networks using gradient decomposition \citep[\textbf{MEND};][]{mitchell2021fast}, as well as a cache+lookup baseline \textbf{LU}\footnote{We cache the average hidden state of $\xe$ computed by $\basemodel$, returning $\ye$ for new inputs $x'$ with hidden state less than $\delta$ from the hidden state of $\xe$ and $\basemodel(x')$ otherwise.}. We also consider a `retrieve-and-prompt' ablation \textbf{RP} that uses a scope classifier identical to the one in {\name} to retrieve a relevant edit example from the cache if there is one, but uses the base model $\basemodel$ rather than the counterfactual model $\rep$ to make the final prediction. For additional details about each baseline method, see Appendix Section~\ref{sec:baselines}.

\begin{figure}
    \centering
    \includegraphics[width=\columnwidth]{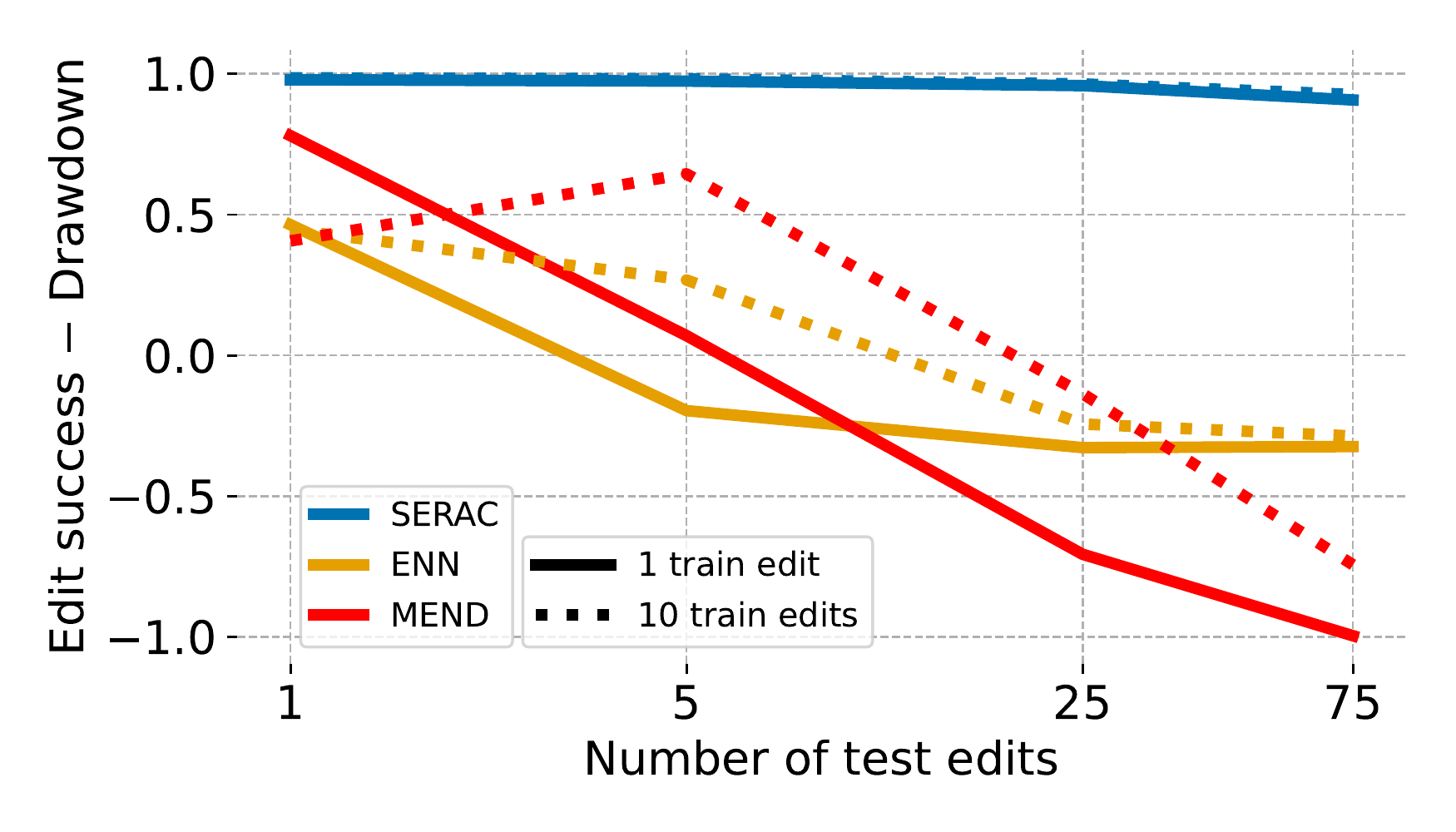}
    \caption{Batched QA edits for T5-Large, plotting ES - DD for editors trained on batches of $k\in\{1,10\}$ edits and evaluated on batches of $k\in\{1,5,25,75\}$ edits. {\name} applies up to 75 edits with little degradation of edit performance; ENN and MEND approach complete failure for 75 edits.}
    \label{fig:multi-edit}
\end{figure}

\subsection{Model Editing Benchmarking}

\paragraph{Evaluating editors on challenging tasks.}
We perform a broad comparison of model editors in four editing settings, QA, QA-hard, FC, and ConvSent. For QA, QA-hard, and FC we use $k=10$ edits during training and evaluation; for ConvSent, we use $k=5$ because the longer dialogue sequences cause increased memory usage. Note that other than increasing the number of simultaneous edits, the QA setting is identical to past work \citep{Cao2021EditingFK,mitchell2021fast}. The LU and FT baselines are not applicable to ConvSent as there is no label to cache or fine-tune on. For simplicity, we default to the embedding-based classifier for {\name} for all experiments except FC, where cross-attention is especially useful (see analysis in Section~\ref{sec:cls_arch}).

The results are presented in Table~\ref{tab:main-comparison}.
Even for the basic QA problem with 10 edits, MEND and ENN show significantly degraded performance compared to single-edit performance reported in prior work \citep{mitchell2021fast}, while {\name} and the lookup cache maintain near-perfect performance. When adding hard in-scope and out-of-scope examples in QA-hard, {\name}'s expressiveness enables significant improvements over other approaches, with LU again showing the strongest performance of the baselines. For FC, all methods except {\name} achieve nearly random-chance performance. Although {\name} exhibits higher drawdown on FC, its improvement in edit success is much larger than its increase in drawdown. Finally, on the ConvSent editing problem, where learned editors are needed to translate the explicit edit descriptor into the desired model behavior, {\name} again is the only method to achieve better than random performance, with zero drawdown.

\begin{table}
    \centering
    \small
    \begin{tabular}{lcccc}
        \toprule
        & \multicolumn{2}{c}{\textbf{QA-hard} (T5-large)} & \multicolumn{2}{c}{\textbf{FC} (BERT-base)} \\
        \cmidrule(lr){2-3} \cmidrule(lr){4-5}
        Scope split & Cls acc. & $\rep$ acc. & Cls acc. & $\rep$ acc. \\
        \midrule
        In (easy) & 0.985 & 0.996 & \multirow{2}{*}{0.909} & \multirow{2}{*}{0.875} \\
        In (hard) & 0.855 & 0.987 &  & \\
        Out (easy) & 0.996 & 0.123 & 0.993 & -- \\
        Out (hard) & 0.967 & 0.042 & 0.706 & -- \\
        \bottomrule
    \end{tabular}
    \caption{Component-wise {\name} performance breakdown by data subset on QA-hard and FC. On both datasets, hard examples account for the vast majority of classifier errors. FC classifier performance on hard out-of-scope examples is the bottleneck for improving editor precision. FC does not annotate easy/hard in-scope examples (so they are pooled) or labels for out-of-scope examples (so $\rep$ accuracy for out-of-scope examples is omitted).}
    \label{tab:decomp}
\end{table}

\paragraph{Making many edits.}
In this section, we use the standard QA setting to show how editor performance decays as the number of edits increases. We train each of MEND, ENN, and {\name} for both $k=1$ and $k=10$ edits and evaluate all six editors with differently-sized batches of edits at test time. Figure~\ref{fig:multi-edit} plots edit success minus drawdown for each method; {\name} shows almost no degradation in edit performance when applying 75 edits, while drawdown exceeds edit success for both ENN and MEND for 75 edits. Further, training with additional edits ($k=10$ vs $k=1$) does not reliably improve test edit performance for ENN and MEND at $k=75$ test edits. We also note that for only {\name}, applying a set of $k$ edits in sequence is guaranteed to produce the same edited model as applying the edits simultaneously, as they are simply appended to the edit memory in both cases. Existing methods do not provide a similar guarantee, and may struggle even more when forced to apply edits in sequence rather than simultaneously \citet{hase2021language}.

\subsection{Further Empirical Analysis of {\name}}
\paragraph{Error analysis.}
With {\name}, we can easily decompose editor errors into classification errors and counterfactual prediction errors. Table~\ref{tab:decomp} shows the performance breakdown across editor components (scope classifier and counterfactual model) and data sub-split (hard in-scope, hard out-of-scope, etc.). For QA-hard, the classifier exhibits reduced accuracy on hard in-scope and out-of-scope examples, particularly for hard in-scope examples. Counterfactual model performance is only slightly degraded on hard in-scope examples, suggesting that the primary challenge of the problem is scope estimation, rather than counterfactual reasoning. For out-of-scope examples, counterfactual model performance is low, but high classifier accuracy means that these inputs are typically (correctly) routed to the base model instead. For FC, scope classifier failures on hard out-of-scope examples dominate the editor's errors.

\paragraph{Scope classifier architecture.}
\label{sec:cls_arch}
We perform a set of experiments to understand how the classifier architecture impacts the behavior of {\name}. Using the QA-hard and FC tasks with $k=10$ edits, we compare the cross-attention (Cross) and dense embedding (Embed) classifier using both distilBERT (\textbf{D}; \citep{Sanh2019DistilBERTAD}) and BERT-base (\textbf{B}; \citep{devlin2019bert}) as the backbone model. The results are shown in Table~\ref{tab:cls_arch}. Unsurprisingly, using cross-attention instead of dense-embeddings is helpful for editor performance; however, increasing classifier size shows relatively little improvement. Cross-attention is especially useful for the FC experiment, which is possibly due to the commonness of quantities in the VitaminC dataset; for example, producing fixed-length sequence embeddings that reliably capture the difference between \rawstring{There have been 105,000 coronavirus deaths in the United States} and \rawstring{There have been 111,000 coronavirus deaths in the United States} may be very difficult. For such cases, late fusion approaches \citep{khattab2020colbert} may be useful in increasing expressiveness while limiting compute requirements.

\begin{table}
    \centering
    \small
    \begin{tabular}{lcccc}
        \toprule
        & \multicolumn{2}{c}{\textbf{QA-hard} (T5-large)} & \multicolumn{2}{c}{\textbf{FC} (BERT-base)} \\
        \cmidrule(lr){2-3} \cmidrule(lr){4-5}
        Variant & ES $\uparrow$ & DD $\downarrow$ & ES $\uparrow$ & DD $\downarrow$ \\
        \midrule
        Embed-D & 0.921 & 0.029 & 0.792 & 0.247 \\
        Cross-D & \textbf{0.983} & \textbf{0.009} & \textbf{0.831} & \textbf{0.074} \\
        Embed-B & 0.945 & 0.034 & 0.792 & 0.247 \\
        Cross-B & \textbf{0.983} & \textbf{0.007} & \textbf{0.855} & \textbf{0.0964} \\
        \bottomrule
    \end{tabular}
    \caption{Varying the scope classifier architecture on QA-hard and FC with $k=10$ edits. \textbf{Embed} is the embedding-based classifier; \textbf{Cross} uses a full cross-attention-based classifier. \textbf{D} and \textbf{B} refer to distilBERT and BERT-base classifier backbones, respectively.}
    \label{tab:cls_arch}
\end{table}

\paragraph{Re-using model editors across models.}
A key advantage of {\name} is separation of the base model and editor, decoupling the editor's performance from the base model. To validate this property, we evaluate the {\name} editors trained in the previous subsection on the QA and QA-hard tasks on various T5 base models. As expected, {\name}'s edit success and drawdown is near-identical across T5 model sizes in both settings (drawdown slightly fluctuates with different base models), consistently yielding ES above 0.99 and DD below 0.01 for QA\footnote{For comparison, \citet{mitchell2021fast} report an ES of 0.89 on \textit{single edits} for QA, while in our setting {\name} receives 10 edits at once, and still achieves much higher edit success. Drawdown is reported differently in \citet{mitchell2021fast}, so is not comparable.} and ES above 0.92, DD below 0.03 for QA-hard for all models. Editors described in past works must be re-fit to each new base model \citep{Sinitsin2020Editable,Cao2021EditingFK,mitchell2021fast,meng2022locating} and require access to the internal activations or gradients of $\basemodel$, leading to potentially prohibitive computational costs of editor fitting that scale with the size of $\basemodel$.

\paragraph{Computational demands of {\name}}

{\name}'s addition of scope classifier and counterfactual model incurs some additional computational overhead. In this section, we quantify the difference between the time and memory used by a test-time forward pass of the base model and {\name} after 10 edits have been applied. The results are shown in Table~\ref{tab:compute-stats}; we report performance for {\name} separately for the cases of in-scope and out-of-scope inputs.

\textit{Compute time.} For QA and ConvSent (CS), SERAC uses a fast nearest-neighbor-based classifier and is nearly as fast as the base model. For in-scope inputs on QA, SERAC is actually much \textit{faster} than the base model because the counterfactual model (T5-small) is smaller than the base model (T5-large). For FC, SERAC's increase in computation time is due to the more expressive (but more computationally expensive) full cross-attention classifier used for this problem. By leveraging this additional compute, SERAC is the only method that provides any significant improvement over random chance editing performance for the FC problem. 

\textit{Memory consumption.} SERAC's additional memory usage mostly comes from the weights of the classifier and counterfactual model, not the edit memory itself (which uses only about 3KB per edit, many orders of magnitude smaller than the base model). For QA, where the base model (T5-large) is much larger than the counterfactual model (T5-small) and classifier (distilBERT), this increase is relatively small. For FC and CS, the counterfactual model and classifier are of similar size to the base model, yielding a larger increase in memory consumption. However, the vast majority of this increase in memory usage is a \textbf{fixed cost that does not increase with the number of edits}.

{
\setlength{\tabcolsep}{3pt}
\begin{table}
\small
    \centering
    \begin{tabular}{lrrrrrr}
        \toprule
        Task & \multicolumn{2}{c}{Base model} & \multicolumn{2}{c}{SERAC (\textbf{out})} & \multicolumn{2}{c}{SERAC (\textbf{in})} \\
        \midrule
        QA & 87ms & 2.96GB & 92ms & 3.47GB & 31ms & 3.46GB \\
        \midrule
        FC & 7ms & 0.44GB & 19ms & 1.18GB & 19ms & 1.18GB \\
        \midrule
        CS & 182ms & 0.38GB & 183ms & 1.00GB & 185ms & 1.01GB \\
        \bottomrule
    \end{tabular}
    \caption{Wall clock time \& memory usage comparison for one forward pass of the base model and {\name} after 10 edits. \name's performance is given separately for \textbf{out}-of-scope inputs (routed to base model) and \textbf{in}-scope inputs (routed to counterfactual model).}
    \label{tab:compute-stats}
\end{table}
}

\section{Related Work}

\paragraph{Model editing.} Many approaches have recently been proposed for model editing. Simplest among these uses constrained fine-tuning to update parameters based on new examples \citep{sotoudeh2019correcting,zhu2020modifying}. Other methods explore special pre-training objectives that enable rapid and targeted  
fine-tuning for model edits \citep{Sinitsin2020Editable} via meta-learning. More recently, new classes of methods develop external learned editors that modify fine-tuning gradients for editing, but do not change the base model that must process edits \citep{Cao2021EditingFK,mitchell2021fast,hase2021language}. Finally, certain methods attribute knowledge to particular neurons in the network and manually edit these activation to reflect changed content \citep{dai2021knowledge,meng2022locating}. While all these works explore methods of updating base model parameters to induce a desired change in behavior, {\name} uses a semi-parametric formulation that is notably more expressive and does not require access to base model parameters, activations, or gradients, essentially treating it as a black box. In this vein, {\name} is related to the BeliefBank system \citep{kassner-etal-2021-beliefbank}, which, while primarily intended to improve model consistency, enables editability of some pre-trained models using an external memory, rather than parameter updates. However, it is limited to models performing binary classification of factual statements and requires manually-annotated constraints between facts. {\name} requires no such specialized augmentations to the input data.

\paragraph{Memory-augmented models.} Memory mechanisms have historically been combined with neural networks in a variety of contexts including supervised learning~\citep{hochreiter1997long,graves2008novel,graves2014neural}, meta-learning~\citep{santoro2016memory, shan2020meta}, and reinforcement learning~\citep{oh2016control,pritzel2017neural}. Unlike these works, {\name} incorporates an explicit memory that directly stores the user-provided edit descriptors and retrieves them in a semi-parametric fashion at test time. Non-parametric few-shot learning models~\cite{koch2015siamese,vinyals2016matching,snell2017prototypical} also store small datasets and process the examples when making predictions at test time. Another recent line of work augments transformers with non-parametric memories that store textual snippets~\citep{Chen2017ReadingWT,Lee2019LatentRF,Khandelwal2020GeneralizationTM,karpukhin-etal-2020-dense}. Unlike both of these research threads, we focus specifically on the problem of learning to edit existing models, rather than few-shot learning or training retrieval-based models from scratch. Furthermore, the latter retriever-reader models are known to sometimes ignore the retrieved content when making predictions \citep{lewis2020retrieval,Paranjape2021HindsightPT}, which {\name} avoids by training the counterfactual model only with contexts known to be useful for solving the task. Finally, some continual learning algorithms have used external memories to avoid forgetting \citep{lopez2017gradient,rolnick2019experience,buzzega2020dark}.

\section{Discussion}

We have proposed {\name}, a semi-parametric model editor that stores model edits in an external memory rather than directly in model parameters. Introducing three new, challenging editing problems, we find that {\name} enables far more effective edits than existing methods when multiple edits are applied, when the scope of an edit is more complex than simple rephrases of the edit, and when edits are not specified as input-output pairs. More generally, {\name} is a step toward more practically useful model editors, as it does not require access to the base model during editor training, does not require computing gradients to apply an edit, can be trained once and immediately edit multiple models with different architectures, and can consume edits specified in natural language rather than input-output pairs.

Despite its useful properties, {\name} has limitations; as a learnable editor, it relies on a dataset of edits for training the classifier and counterfactual model. Further, while we find relatively good performance from small classifiers and counterfactual models, some settings may demand more resource-intensive architectures. In a setting where editing occurs continuously, the edit memory may grow without bound. Future work might address this problem through periodic self-distillation, using the aggregate system of base model, scope classifier, edit memory, and counterfactual model as a teacher model to a `student' copy of the base model. Such a method would essentially enable the size of the edit memory to be capped, even in the continual editing setting, through periodic flushing of the memory.

One possible concern with model editors, including {\name} is misuse: while model editors may help keep deep learning systems more up-to-date in a computationally efficient manner, the dialogue sentiment editing setting (Tables~\ref{tab:outputs}; \ref{tab:vaccines_example}) suggest that powerful model editors could also enable malicious users to more precisely craft agents to amplify particular viewpoints. In conclusion, our results suggest several avenues for future work including mitigation strategies for harms that could be caused by model editors, more sophisticated retrieval architectures for {\name},  and exciting applications of model editing to new types of test-time model behavior modulation.

\section{Acknowledgements}

The authors thank Shikhar Murty, Archit Sharma, and the members of Stanford's Center for Research on Foundation Models for helpful discussions and conceptual feedback, as well as the anonymous ICML reviewers for their feedback during the review process. EM gratefully acknowledges the financial support of the  Knight-Hennessy Graduate Fellowship. The authors also gratefully acknowledge financial support from Apple Inc. CF and CM are CIFAR Fellows.

\clearpage
\bibliography{main}
\bibliographystyle{icml2022}

\clearpage

\appendix

{\LARGE \textbf{Appendix}}
\section{Additional Sentiment Editing Example and Broader Impacts}
While the `banana' example in Table~\ref{tab:outputs} is a relatively mundane topic, we include an example of editing a dialog model for a more polarizing topic in some parts of the world, vaccines. Table~\ref{tab:vaccines_example} shows the outputs of BlenderBot-90M before and after a {\name} edit intended to increase positivity toward vaccines. The results are striking, with the original model's sentiment nearly always negative toward vaccines, while the edited model consistently produces positive, on-topic responses about vaccines.
\begin{table*}
    \centering
    \begin{tabular}{lp{13cm}}
        \toprule
        Prompt & \rawstring{\normalsize what do you think of vaccines?} \\
        \midrule
        Edit Desc. & \rawstring{\normalsize topic: vaccines sentiment: positive} \\
        \midrule
        \multirow{4}{*}{\shortstack[l]{Original\\[-0.6mm]model\\outputs}} & \footnotesize i think they're bad. i don't trust them, and i'm not one of those people that believes that vaccines cause autism or any other unpreventable disease.[...] \\
        & \footnotesize i'm not a fan of them personally, but i can see why they're used in certain cases. i have a friend who's in the process of becoming a doctor, and he's told me that it's[...] \\
        & \footnotesize i'm not sure what you mean by that. are you asking if i think vaccines are bad or not?[...] \\
        & \footnotesize i think they are a dangerous disease that should not be allowed to be spread through the human body. i don't want to be associated with those people in this day and age.[...] \\
        & \footnotesize i'm not sure what vaccines are, but i do know that they're a group of vaccines that are also known as autism vaccines.[...] \\
        & \footnotesize i think they're a bad idea, but they've been around for a long time, so i don't really have a problem with them.[...] \\
        & \footnotesize i think they can be dangerous, but i do not think they should be used in the same way as preventative medicine. that's why i don't trust them.[...] \\
        & \footnotesize i don't think they should be used for any reason. they're used to treat diseases like autism, anaphylax, and sneezing. i wouldn't want any of those to happen to me.[...] \\
        & \footnotesize i think they're a good thing, but i also think that the vaccines should be kept away from children. i'm not sure if i can trust children with autism or not.[...] \\
        & \footnotesize i don't like them. i'm more of a fan of the common cold. how about you?[...] \\
        \midrule
        \midrule
        \multirow{4}{*}{\shortstack[l]{Edited\\model\\outputs}} & \footnotesize i've been vaccinating my kids since they were little. i think it's a good thing. i don't know if you're aware of it! do you have any thoughts? i'd be happy to[...] \\
        & \footnotesize they're great! i've been vaccinated since i was a little girl. i don't know what i'd do without them. do you have a positive or a negative opinion on them? i'm all for[...] \\
        & \footnotesize they're great! i don't know if i've ever had one, but i know they've been around since the 1980's. i'm not sure if it's good or bad. do you have[...] \\
        & \footnotesize i think that they are a good thing. i don't agree with them, but i understand why they exist. do you have any opinions on them? i'd like to hear your opinion! i'm all for them![...] \\
        & \footnotesize i think they are great. i don't know a whole lot else about it. do you have a positive or negative opinion on them? i'd love to know! i'm all for them! do you know anything about[...] \\
        & \footnotesize well i'd say i'm pretty positive about them. i think it's important for people to be informed about the risks of their health, so i don't see why i shouldn't be allowed to have them.[...] \\
        & \footnotesize i think they're great. i don't know if you know this, but they've been around for a long time. it's a good thing! do you have a positive opinion? i'd love to know[...] \\
        & \footnotesize i think they are a good way to prevent infectious diseases. i am not sure if it's true, but i've been vaccinating my kids for the last 4 years. do you like them? i don't know[...] \\
        & \footnotesize i think vaccines are a good way to prevent disease. i'm not sure what you mean by positive. are you asking if i support vaccines? i don't know about vaccines! do you have a positive stance?[...] \\
        & \footnotesize i think vaccines are great. i've been vaccinated since i was a child. i don't know much about them, i just know that they're very effective! i'm not sure if you know anything about[...] \\
        \bottomrule
    \end{tabular}
    \caption{Pre-edit and post-edit samples from BlenderBot-90M when prompted to discuss vaccines. Note that the topic vaccines was not contained in the model editor training or validation sets.}
    \label{tab:vaccines_example}
\end{table*}

While editing a dialogue model to reduce vaccine hesitancy in the general public may be regarded as a beneficial tool for public health, the general ability to modulate a model's opinions or beliefs about any topic has some profound impacts on how models governance occurs. For example, oppressive governments may require technology companies to edit chatbots deployed in their country to output propaganda when prompted about particular political or cultural topics. Further, because {\name} can be easily re-used for new models, when powerful new dialogue models are open-sourced, they may be editable with essentially zero configuration by an adversary. Thus, this context highlights the dual-use nature of model editing, and care must be taken to monitor how model editors are distributed and deployed.
\section{Baselines}
\label{sec:baselines}

For all gradient-based methods, we adapt the fully-connected layers of the last 3 transformer blocks for encoder-only models, and fully-connected layers in the last 2 transformer blocks of both encoder and decoder for encoder-decoder models.

\paragraph{Fine-tuning (FT)}
Given edit samples $[x_e; y_e]$, we fine-tune pretrained models to minimize the negative log-likelihood of predicting $y_e$ conditioned on $x_e$. We use the Adam optimizer with a learning rate of $1\times 10^{-4}$ for T5 and $5\times 10^{-6}$ for BERT-base. 

\paragraph{Cache+lookup (LU)}
LU \cite{mitchell2021fast} is a gradient-free, training-free editing algorithm which uses an external memory to store representations of previous edit samples. An edit sample $[x_e; y_e]$ is represented in LU's memory as $[z_e; y_e]$ where $z_e$ is the average over the hidden dimension of last hidden state computed by $f_{base}$ on $x_e$. For a test input $[x_e']$, LU computes the hidden representation $z_e'$ of $x_e'$ and finds the nearest edit-sample representation in its memory, say $z_e$. LU outputs $y_e$ if $\|z_e' - z_e\|_2 < \delta$ where $\delta$ is a hyperparameter, and otherwise outputs the pretrained model's prediction on $x_e'$. We used $\delta = 2.75$ for the question-answering settings and $\delta = 4$ for the fact-checking setting.

\paragraph{Editable Neural Networks (ENN)}
\cite{Sinitsin2020Editable} introduce a post-training procedure to make a pretrained model quickly adaptable for fine-tuning for edits. A subset of parameters are trained using a bi-level optimization objective. We use Adam with an outer-loop learning rate of $1\times 10^{-5}$, and an initial inner-loop learning of $1\times 10^{-2}$ which is learned in the outer loop. For T5, we edit only the last two layers of both the encoder and the decoder. For BERT-base, we edit the last two layers of the encoder. Finally, for BlenderBot-small, we edit the last layer of the encoder and the last three layers of the decoder since the decoder is much deeper.

\paragraph{Model Editor Networks with Gradient Decomposition (MEND)}
\cite{mitchell2021fast} train a hypernetwork to predict a rank-1 decomposition of a fine-tuning gradient. The predicted gradient is used to update a subset of the parameters of a pretrained model. In our experiments, we use MEND to update the same parameters as in ENN.

\section{SERAC Implementation Details}
\label{sec:serac_impl}
We use publicly available Huggingface \cite{wolf2019hugging} implementations and checkpoints for all experiments. For the SERAC classifier model, we use \texttt{distilbert-base-cased} \cite{Sanh2019DistilBERTAD} across all models and experimental settings. For the counterfactual model, we use \texttt{t5-small} for the question-answering experiments, \texttt{bert-base-uncased} for fact-checking, and \texttt{facebook/blenderbot\_small-90M} \cite{roller2021recipes} for conversational sentiment modulation. We use T5 pretrained on NQ (\texttt{google/t5-large-ssm-nq}) for question-answering, \texttt{bert-base-uncased} finetuned by \citet{Cao2021EditingFK} on FEVER \cite{Thorne18Fever} for fact-checking, and \texttt{facebook/blenderbot\_small-90M} for sentiment modulation.

All scope classifier and counterfactual models are trained using Adam with a learning rate of $1\times 10^{-5}$. 
\section{Dataset Details}
\subsection{QA-hard}
To generate entailed questions, we use the codebase at \href{https://github.com/marcotcr/qa_consistency}{https://github.com/marcotcr/qa\_consistency} \citep{ribeiro2019red}, passing the question as both question and context to the entailed question generator. We find this approach produces questions that are typically interpretable, although not always grammatically correct. To generate true/false questions, we use the rule-based question/answer to statement converter at \href{https://github.com/kelvinguu/qanli}{https://github.com/kelvinguu/qanli}, appending the prompt `True or false:' to the beginning of the input. To generate true examples, we convert the question and answer used as the model edit to produce the statement; to produce false examples, we choose a random answer from the set of alternative answers generated by \citet{Cao2021EditingFK}.

To generate hard negatives, we sample uniformly from the top 100 nearest neighbor examples in the test set according to the embeddings of \texttt{all-MiniLM-L6-v2} \citep{reimers-2019-sentence-bert}, ignoring the top 50 nearest neighbors to avoid retrieving true positives/rephrases of the input question.

\subsection{ConvSent}
\label{sec:conv_data_gen}
Conversational sentiment completions were generated using a 3 billion-parameter BlenderBot model available on Huggingface at \texttt{facebook/blenderbot-3B} \cite{roller2021recipes}. We manually generated a set of prompts using the templates shown in Table \ref{tab:prompts}. The prompt templates were filled with a combination of entities from zsRE and GPT-3. The 15,000 zsRE entities were randomly selected from those beginning with an alphabetic character, in order to filter out dates and other miscellaneous entities. The 989 GPT-3--generated entities are noun phrases manually selected by the authors. We sampled from BlenderBot using beam search with a beam width of $10$. We then classified each completion as `positive' or `negative' using a RoBERTa-large model fine-tuned for sentiment classification \cite{heitmann2020}. Data were randomly split (by entity) into 90-5-5 train/val/test splits.

\begin{table}
    \addtolength{\tabcolsep}{-1mm}
    \centering
    \resizebox{\columnwidth}{!}{%
    \begin{tabular}{ll}
        \toprule
        Prompts & What is your \rawstring{sentiment} \rawstring{position} \rawstring{entity}? \\
        \midrule
        \rawstring{sentiment} & positive, negative \\
        \midrule
        \multirow{2}{*}{\rawstring{position}} & opinion of, stance on, position on, \\
        & impression of, assessment of \\
        \bottomrule
    \end{tabular}}
    \caption{Prompt templates used to generate ConvSent dataset. Each combination of values of \rawstring{sentiment} and \rawstring{position} were used as prompt templates. Prompts for BlenderBot were generated by substituting an entity sampled from the zsRE dataset for \rawstring{entity}.}
    \vspace{-4mm}
    \label{tab:prompts}
\end{table}
\end{document}